\documentclass[11pt]{article}

\usepackage[T1]{fontenc}
\usepackage[utf8]{inputenc}
\usepackage[a4paper,margin=28mm]{geometry}
\usepackage{graphicx}
\usepackage{natbib}
\usepackage{amsmath}
\usepackage{authblk}
\usepackage{microtype}
\usepackage[bottom]{footmisc}
\usepackage{caption}
\usepackage{subcaption}
\usepackage{newtxtext}
\usepackage[varvw]{newtxmath}
\usepackage{booktabs}
\usepackage{tabularx}
\usepackage{xurl}
\usepackage[hidelinks]{hyperref}

\begin{document}

\title{Machine Learning in Fish Farming}
\author[1]{Fearghal O'Donncha}
\author[2]{Nikos Papandroulakis}
\author[3]{Jennie Korus}
\author[4]{Abigail Langbridge}
\author[4]{Alexander Timms}
\author[5]{Konstantinos Topouzelis}
\author[6]{Abdul Baseer Khan}
\author[6]{Shree Rama Kamal Kumar Vegu}
\author[6]{Mahtab Sarvmaili}
\author[7]{Ryan Mowat}
\author[8]{Rhanna Turberville}
\author[3]{Tyler Sclodnick}
\author[6,9]{Christopher Whidden}
\affil[1]{IBM Research Europe, Dublin, Ireland}
\affil[2]{Hellenic Centre for Marine Research, Crete, Greece}
\affil[3]{Innovasea Marine Systems Canada, Nova Scotia, Canada}
\affil[4]{Imperial College London, United Kingdom}
\affil[5]{University of the Aegean, Greece}
\affil[6]{Dalhousie University, Halifax, Nova Scotia, Canada}
\affil[7]{RS Aqua, Portsmouth, United Kingdom}
\affil[8]{Scottish Sea Farms, Orkney, United Kingdom}
\affil[9]{DeepSense, Nova Scotia, Canada}
\date{}

\maketitle

\begin{quote}
\small\textbf{Preprint notice.} This is a preprint of the following chapter: Fearghal O'Donncha, Nikos Papandroulakis, Jennie Korus, Abigail Langbridge, Alexander Timms, Konstantinos Topouzelis, Abdul Baseer Khan, Shree Rama Kamal Kumar Vegu, Mahtab Sarvmaili, Ryan Mowat, Rhanna Turberville, Tyler Sclodnick and Christopher Whidden, ``Machine Learning in Fish Farming,'' published in \emph{Digital Twin for Marine Fish Farming}, edited by Fearghal O'Donncha, Martin F\o re and Jon Grant, 2026, Springer, reproduced with permission of Springer. The final authenticated version is available online at: \url{https://doi.org/10.1007/978-3-032-10362-8_3}.
\end{quote}

\begin{abstract}
This chapter explores how machine learning (ML) is transforming aquaculture, with a particular focus on enhancing decision-making processes and improving operational efficiency. The chapter is structured to first introduce the challenges in aquaculture and the role of AI, then provide an overview of ML techniques in the context of aquaculture, followed by applications, emerging trends, future directions, and case studies. The focus is on real-world applications of ML techniques, including Random Forest, Convolutional Neural Networks (CNNs), and Recurrent Neural Networks (RNNs), as well as emerging technologies such as Graph Neural Networks (GNNs) and transformer-based models. Key applications include biomass estimation, species recognition, behavioural analysis, and environmental forecasting. The chapter also highlights the synergy between ML and the Internet of Things (IoT) for real-time monitoring and decision support. Ultimately, ML-driven innovations have the potential to revolutionize fish farming, leading to more efficient, sustainable, and productive practices in the aquaculture industry.
\end{abstract}


\section{Introduction}
\label{sec:intro}
Aquaculture has emerged as a critical component of global food security, supplying nearly 50\% of the world's fish consumption. It is recognized as one of the fastest-growing food production sectors, contributing significantly to the supply of high-quality protein, particularly in regions where access to such food sources is limited \citep{naylor202120, bene2016contribution}. As global demand for seafood continues to rise, the expansion of aquaculture is essential to meet these needs sustainably. However, the industry faces multiple challenges that threaten its long-term viability.

\subsection{Challenges in Aquaculture}

Despite its rapid growth, aquaculture is constrained by several critical challenges that affect sustainability, efficiency, and profitability.
\begin{itemize}
    \item \textbf{Environmental Concerns:} The intensification of aquaculture can contribute to habitat degradation, water pollution, and excessive nutrient loading, leading to issues such as harmful algal blooms and oxygen depletion \citep{godfray2010food}.
    
    \item \textbf{Disease Management:} Infectious diseases pose a significant threat to fish health, resulting in high economic losses. Disease outbreaks require improved early detection and intervention strategies to prevent large-scale losses \citep{mandal2023role}.
    
    \item \textbf{Feeding Efficiency and Resource Utilization:} Feed is one of the most expensive inputs in aquaculture, and inefficient feeding practices can lead to increased waste and production costs . Optimizing feed conversion ratios is essential to reduce environmental impact and improve profitability \citep{cooney2021designing}.
    
    \item \textbf{Labour Shortages and Supply Chain Vulnerabilities:} The COVID-19 pandemic exposed weaknesses in aquaculture operations, highlighting issues such as workforce shortages and logistical disruptions that can impact production efficiency \citep{love2021emerging}.
\end{itemize}

These challenges underscore the urgent need for greater automation and data-driven approaches to ensure efficient, sustainable, and resilient aquaculture operations \citep{klinger2012searching}.

\subsection{The Role of Artificial Intelligence in Aquaculture}

Artificial Intelligence (AI) is revolutionizing aquaculture by providing advanced analytics, automation, and predictive capabilities. Machine learning (ML), a subset of AI, enables data-driven decision-making and enhances key aspects of aquaculture management, including:

\begin{itemize}
    \item \textbf{Optimized Feeding Strategies:} AI-driven analysis of fish behaviour and biomass estimation allows for precise feeding schedules, minimizing waste and improving growth rates.
    
    \item \textbf{Early Disease Detection:} ML models analyse behavioural patterns, water quality data, and historical trends to detect early signs of disease outbreaks, allowing for timely interventions.
    
    \item \textbf{Environmental Monitoring and Prediction:} Predictive models improve water quality management by forecasting dissolved oxygen levels, temperature fluctuations, and harmful algal blooms \citep{odonncha2019precision}.
    
    \item \textbf{Automation of Farm Operations:} AI-driven IoT systems enable automated monitoring of fish health, water conditions, and infrastructure maintenance, reducing reliance on manual labour \citep{mandal2023role}.
\end{itemize}

The integration of AI into aquaculture is not merely a technological advancement but a necessary step toward increasing efficiency and ensuring long-term sustainability \citep{maksimovic2018greening}.

\subsection{Scope of This Chapter}

This chapter explores the role of machine learning in aquaculture, focusing on its applications in farm management, decision-making, and operational efficiency. Given the complexity of aquaculture, which intersects with fields such as oceanography, environmental science, industrial automation, and logistics, the discussion is centered on how AI-driven approaches can enhance on-farm operations. The key question addressed is: \textit{How can the combination of human expertise and AI-based insights lead to more effective and sustainable aquaculture practices?}

The chapter is structured as follows:

\begin{itemize}
    \item \textbf{Section 2} provides an overview of machine learning techniques relevant to aquaculture.
    \item \textbf{Section 3} discusses key applications, including computer vision for fish monitoring, IoT-based environmental tracking, and predictive modelling for disease detection and feeding efficiency.
    \item \textbf{Section 4} examines emerging trends, such as Graph Neural Networks (GNNs), large language models, and AI-integrated digital twins for aquaculture.
    \item \textbf{Section 5} presents case studies illustrating real-world AI applications in aquaculture.
    \item \textbf{Section 6} provides conclusions and outlines directions for future research.
\end{itemize}

This structured approach ensures a comprehensive exploration of machine learning in aquaculture, highlighting its potential to drive innovation, efficiency, and sustainability in the industry.

\section{Machine Learning}
\label{sec:ML}

Research into applications of artificial intelligence (AI) and machine learning (ML) in aquaculture and other maritime applications has accelerated at a rapid pace over the last 5 years or so. AI is a broad field focused on creating smart machines capable of performing tasks that require human intelligence. ML, on the other hand, is a subset of AI that specifically involves developing algorithms that allow computers to learn and make decisions from data. In essence, while AI encompasses a wide range of technologies mimicking human abilities, ML is a specific approach within AI that focuses on data-driven learning and prediction.  

Given access to sufficient and high-quality data, ML techniques can detect, quantify, and predict complex phenomena relevant to aquaculture operations.
While physics-based modelling involves providing a set of inputs to a model which generates the corresponding outputs based on a non-linear mapping encoded from a set of governing equations, supervised machine learning instead learns the requisite mapping by  being shown large number of corresponding inputs and outputs. 
However, this data-driven approach is only as effective as the data it relies on—a principle often summarized as \textit{rubbish in, rubbish out}. If the training data is noisy, sparse, biased, or poorly labelled, the resulting model will inevitably reflect those flaws, regardless of how sophisticated the algorithm is.
Thus, the selection of an appropriate ML model and its hyperparameters—external settings not learned from data but critical to model behaviour—must be matched with careful data curation and preprocessing. Parameters internal to the model are optimized during training, but no model can overcome fundamentally flawed input. Data quality, therefore, is not a peripheral concern but a central determinant of model success.

The remarkable growth and success of machine learning in recent years has been largely driven by three pivotal factors: the exponential increase in data volumes, significant advancements in computational power, and substantial innovations in machine learning algorithms. The proliferation of data from myriad sources such as sensors, Internet of Things (IoT) devices, and social media has provided an unprecedented amount of raw material for machine learning models. This abundance of data enables researchers and practitioners to train more robust and accurate models, particularly in domains like natural language processing, image recognition, and predictive analytics.

Concurrently, there has been a dramatic enhancement in computational power, primarily due to the advent of specialised hardware such as Graphics Processing Units (GPUs) and Tensor Processing Units (TPUs). These advancements have drastically reduced the time required for training complex ML models, making it feasible to process large datasets and perform more sophisticated computations. Moreover, the evolution of machine learning algorithms, especially in the realms of deep learning and reinforcement learning, has opened new horizons. Techniques such as convolutional neural networks (CNNs) \citep{oshea2015introduction}, recurrent neural networks (RNNs) \citep{schmidt2019recurrent}, and transformers \citep{vaswani2017attention}, have set new standards in accuracy and efficiency, pushing the boundaries of what was previously considered achievable in ML. These developments have collectively propelled the field into a new era of innovation and application, transforming theoretical concepts into practical solutions across various industries.

\subsection{Machine learning algorithms}
\label{sec:ML_algos}
Machine learning algorithms can be classified based on various criteria, each providing a different perspective on how the algorithms function and are applied. Some of the primary ways to categorise these algorithms include:
\begin{enumerate}
    \item \textbf{Based on Learning Style:}
    \begin{itemize}
        \item \textbf{Supervised Learning:} Algorithms that learn from labelled training data, aiming to make predictions or decisions. Examples include linear regression, logistic regression, support vector machines, and neural networks.
        \item \textbf{Unsupervised Learning:} Algorithms that work with unlabelled data to find underlying patterns or structures. Common examples are k-means clustering, hierarchical clustering, and principal component analysis.
        \item \textbf{Semi-Supervised Learning:} These algorithms use a mix of labelled and unlabelled data, which are useful when acquiring a fully labelled dataset is expensive or impractical.
        \item \textbf{Reinforcement Learning:} Algorithms that learn to make decisions by performing actions in an environment to achieve a goal. These algorithms learn from the feedback of their own actions rather than from explicit teaching \citep{madondo2023swat}.
    \end{itemize}
    \item \textbf{Based on Functionality:}
    \begin{itemize}
        \item \textbf{Classification Algorithms:} Used for tasks where the output is a discrete label. For instance, identifying whether a fish is feeding or not.
        \item \textbf{Regression Algorithms:} Used for predicting a continuous-valued output, such as the dissolved oxygen conditions in a cage over time.
        \item \textbf{Clustering Algorithms:} Aimed at grouping sets of similar items in data, such as grouping fish based on health status.
        \item \textbf{Dimensionality Reduction Algorithms:} Used to reduce the number of features in a dataset, either for compression or improved efficiency in subsequent tasks.
    \end{itemize}
    \item  \textbf{Based on Approach or Technique:}
        \begin{itemize}
            \item \textbf{Decision Trees:} Algorithms that model decisions and possible consequences, including chance event outcomes, resource costs, and utility.
            \item \textbf{Bayesian Algorithms:} Based on Bayes’ Theorem, these algorithms are particularly used for classification and prediction.
            \item \textbf{Bagging:} Techniques like Random Forest that combine the results of multiple models to improve the overall performance.
            \item \textbf{Boosting:} Methods like AdaBoost and XGBoost that sequentially build models, with each model trying to correct the errors of its predecessor.
            \item \textbf{Deep Learning:} A subset of machine learning based on artificial neural networks with representation learning. Deep learning can involve a high level of automation in predictive modelling.
        \end{itemize}
    \item \textbf{Based on Type of Data or Application:}
    \begin{itemize}
        \item \textbf{Text Analysis/Natural Language Processing (NLP):} Algorithms specialised in processing and analysing text data, like sentiment analysis or topic modelling.
        \item \textbf{Time Series Analysis:} Algorithms that are designed to analyse time-ordered data, useful in fields like environmental science, finance, and economics.
    \end{itemize}
\end{enumerate}
Each classification offers a different lens through which to understand and apply machine learning algorithms, often depending on the nature of the data and the specific problem at hand. It's essential for practitioners to comprehend the various strengths and limitations of different algorithms to create an efficient experimental framework. Although deep learning has garnered widespread success in numerous fields, it's important to note that these approaches often require substantial amounts of data, which may not be feasible for all scenarios. Additionally, in line with the principle of Occam's Razor, it's advisable to gravitate towards the simplest approach that adequately fulfils the requirements of the task.

\subsection{Statistical and Machine Learning}
\label{subsec:stat_ML}
Methods such as linear regressions, Random Forest, and XGBoost are well-established, robust, and efficient methods commonly used in aquaculture. 
They are appealing due to their simplicity and computational efficiency but also due to their ease of interpretability which is particularly important for industry applications.

\subsubsection{Feature Engineering}
Feature engineering is a critical step in the process of developing machine learning models. It involves selecting, modifying, or creating new features from the raw data to improve the performance of machine learning algorithms \citep{guyon2003introduction}. The essence of feature engineering lies in transforming raw data into a format that is more suitable and effective for regression models, such as linear regression or support vector regression. This process not only enhances the model's ability to uncover underlying patterns in the data but also improves its predictive accuracy.

In the context of regression tasks, feature engineering can involve a variety of techniques. One common approach is feature selection, where irrelevant or redundant features are removed to reduce model complexity and avoid overfitting \citep{blum1997selection}. Another technique is feature transformation, which includes normalisation or standardisation of features to bring them onto a similar scale, or applying mathematical transformations to change the distribution of the features. Moreover, feature construction is often employed, where new features are created from existing ones, either through domain-specific knowledge or by detecting interactions between features. This can be particularly useful in linear regression, where non-linear relationships can be linearised through transformations, allowing the linear model to capture more complex patterns in the data \citep{friedman2001elements}. These engineered features are then used to train the regression model, which aims to predict a continuous outcome based on the input features, thereby providing more accurate and reliable predictions.

\subsection{Deep Learning}

Deep learning (DL), a specialised subfield within ML, enables the extraction of complex patterns from large datasets, representing a major leap in AI capability. While machine learning employs algorithms to parse data, learn from that data, and make informed decisions based on what it has learned, deep learning takes this process a step further. It teaches the computer to perform tasks that come naturally to humans: learning by example. Deep learning is a type of machine learning that trains a computer to perform human-like tasks, such as recognising speech, identifying images, or making predictions. Instead of organising data to run through predefined equations, deep learning sets up basic parameters about the data and trains the computer to learn on its own by recognising patterns using layers of processing \citep{lecun2015deep}.

The key difference between deep learning and traditional machine learning lies in how the data is presented to the system. In traditional machine learning, the algorithm requires structured data to learn and make decisions. This involves a significant amount of manual effort to label and classify data before it’s fed into the model. Deep learning, on the other hand, learns from raw data, employing layers of neural networks to understand it. These neural networks, which are algorithms modelled after the human brain, work by passing data through interconnected layers of nodes, or “neurons”. Each layer extracts a specific feature from the data, building a hierarchy of learned features. This hierarchical feature learning enables deep learning models to tackle complex tasks with high accuracy, making them particularly effective for areas like image and speech recognition \citep{goodfellow2016deep}.

DL can be categorised into subfields based on the key architectures that are typically used:
\begin{enumerate}
    \item \textbf{Convolutional Neural Networks (CNNs):} Specialised for processing data with a grid-like topology, such as images. Used in image recognition, video analysis, and natural language processing \citep{lecun1998gradient}.
    \item \textbf{Recurrent Neural Networks (RNNs):} Designed to work with sequence data, like time series or text. They have the ability to retain information over time, making them suitable for tasks like language modeling and speech recognition \citep{hochreiter1997long}.
    \item \textbf{Long Short-Term Memory Networks (LSTMs):} A type of RNN that is better at capturing long-term dependencies, thus avoiding the vanishing gradient problem common in traditional RNNs.
    \item \textbf{Autoencoders:} Used for unsupervised learning tasks, particularly in dimensionality reduction and feature learning. Autoencoders learn to compress data and then reconstruct it back \citep{hinton2006reducing}. 
    \item \textbf{Generative Adversarial Networks (GANs):} Consist of two networks, a generator and a discriminator, which are trained simultaneously. GANs are used for generating new data that resembles the training data, useful in image generation, super-resolution, and more \citep{goodfellow2020generative}.
    \item \textbf{Transformers}: Initially developed for natural language processing (NLP), transformers have emerged as a versatile and powerful architecture for handling sequential data. Unlike recurrent models, transformers rely on a self-attention mechanism that enables them to process entire sequences in parallel, significantly improving computational efficiency and scalability \citep{vaswani2017attention}. This architectural shift has driven breakthroughs in tasks such as machine translation, sentiment analysis, and text generation. More recently, transformers have been adapted for computer vision and time-series modelling, further broadening their applicability. Transformers underpin today’s large language models (LLMs), playing a pivotal role in the development of generative AI systems now being explored across aquaculture and environmental domains.

\end{enumerate}
The different architectures in deep learning are significant because their design and functionality make it particularly suited for certain tasks, leading to more effective and efficient learning outcomes in diverse applications. 

These architectures can also be combined to leverage advantages of multiple methods. Combining Convolutional Neural Networks (CNNs) with Long Short-Term Memory networks (LSTMs) harnesses the strengths of both architectures, creating a powerful tool for tasks involving both spatial and temporal data. CNNs are adept at extracting spatial features and hierarchical patterns in data, particularly useful in image and video analysis. When this spatial information is extracted by a CNN, it can then be fed into an LSTM, which is designed to handle sequential data and capture temporal dependencies. This combination is particularly effective in video processing tasks, where it's crucial to understand not just individual frames (spatial information) but also the sequence and evolution of these frames over time (temporal information) \citep{donahue2015long}.
In environmental applications, CNNs excel in analysing spatial patterns, such as those observed in satellite imagery of ocean surfaces, while LSTMs contribute by interpreting temporal changes and patterns over time, crucial for understanding dynamic ocean processes like current movements, sea surface temperature changes, and algal blooms. This combination enables a more comprehensive analysis of ocean environments, providing valuable insights for environmental monitoring and climate studies \citep{chen2019hybrid}.

\subsubsection{Deep Learning Summary and Considerations}
Deep learning, with its advanced neural network architectures, offers immense potential for both environmental and industry applications, significantly transforming how we approach complex problems in these domains. For aquaculture, deep learning techniques like CNN and RNN can process vast datasets, such as satellite imagery, IoT data, and weather to extract pertinent insights for management, conservation, and mitigation. Similarly, in operational applications, deep learning can optimise processes in manufacturing, logistics, and predictive maintenance, enhancing efficiency and reducing costs \citep{atitallah2020leveraging}.

However, effectively leveraging deep learning across the myriad aspects of aquaculture necessitates careful consideration of several key factors. The `data appetite' of deep learning models is substantial, requiring large and diverse datasets for training to achieve high accuracy, which are often difficult to collect in industry settings \citep{goodfellow2016deep}. Overfitting is another critical issue, where models might perform well on training data but fail to generalise to new, unseen data, necessitating techniques like regularisation and cross-validation to mitigate this risk. Computational requirements for training sophisticated deep learning models are also significant, often requiring specialised hardware like GPUs, which can be a barrier in terms of resources and cost. 

Lastly, the challenge of interpretability in deep learning models, due to their `black box' nature, can be a significant hurdle, especially in sectors where understanding the decision-making process is crucial for trust and compliance \citep{samek2017explainable}. Thus, while deep learning presents exciting opportunities in environmental and operational sectors, its success hinges on judicious data management, mitigating overfitting, planning for computational resources, and advancing efforts in model interpretability.

\section{Applications of Machine Learning in Aquaculture}

The multifaceted nature of aquaculture, encompassing aspects such as environmental monitoring, disease management, feeding efficiency, and yield optimization, makes it an ideal candidate for the integration of ML techniques. The incorporation of robust ML methodologies not only enhances the accuracy and efficiency of operations but also contributes to sustainable and economically viable aquaculture practices. The adaptability of ML to the diverse data sets generated in aquaculture, from IoT sensor data to satellite imagery, underscores its potential as a key driver of innovation and productivity in the sector.

In a recent review by \cite{zhao2021application}, authors discussed several applications of machine learning in aquaculture:
\begin{enumerate}
    \item \textbf{Biomass Detection of Fish:} Machine learning is used for estimating the size, weight, and count of fish, which is crucial for managing fish populations and predicting yields.
    \item \textbf{Recognition and Classification of Fish:} Techniques are employed for identifying fish species, age, and sex, using machine learning algorithms. This is important for species management and scientific research.
    \item \textbf{Behavioural Analysis:} Machine learning aids in observing and interpreting fish behaviours such as feeding and group dynamics, which is vital for welfare and environmental assessments.
    \item \textbf{Water Quality Parameter Prediction:} Machine learning algorithms are used to predict and monitor various water quality parameters, ensuring optimal conditions for aquaculture.
\end{enumerate}
These examples illustrate ML's growing role in driving more efficient, sustainable, and productive aquaculture practices. The subsequent sections of this chapter will delve further into the diverse applications of machine learning in fish farming, exploring its various implementations and impacts in greater detail.

\subsection{Computer Vision in Aquaculture}

A significant share of research focuses on using computer vision technology to improve observation, detection, and management of caged fish. Traditional methods using manual sampling often harm fish and rely on human judgment, making them unsuitable for modern aquaculture's demands \citep{jeong2020appropriate}. The integration of machine learning and machine vision presents a new, intelligent approach for estimating fish size, weight, numbers, and other biological information more accurately.

The inherent nature of farming fish underwater makes computer vision applications in aquaculture challenging. \cite{yang2021computer} summarised some of these challenges:
\begin{itemize}
    \item \textbf{Colour Similarity:} Fish often change colour to match their environment, leading to a resemblance between the fish and its background. This similarity complicates fine-grained fish classification and accurate segmentation, impacting detection performance.
    \item \textbf{Deformation:} Fish are non-rigid and change posture, causing appearance alterations that can disrupt tracking.
    \item \textbf{Scale Change:} The distance between the camera and fish affects the fish's size in images, influencing detection and counting accuracy. Smaller fish scales include more background, causing errors, while larger scales may miss multiple targets in a single image, leading to tracking and counting issues. Scale adaptation is essential.
    \item \textbf{Occlusion and Disappearance:} Fish may get obscured or disappear temporarily during movement, resulting in tracking failures and loss of continuity in subsequent frames.
    \item \textbf{Illumination Change:} Varying light conditions, from sunrise to sunset and on bright to cloudy days, affect light refraction in water, impacting video/image quality.
    \item \textbf{Low Resolution:} Camera optical and electronic systems can introduce noise and distortions, leading to lower resolution images.
    \item \textbf{Complex Background:} Backgrounds with multiple non-fish objects similar to fish reduce detection accuracy.
    \item \textbf{Weather Change:} Sudden weather shifts, like cloudiness, storms, or typhoons, worsen image contrast, complicating video analysis and target detection.
    \item \textbf{High-Density Scene:} In dense aquaculture environments, severe occlusion during feeding complicates individual fish tracking and behavioral analysis, especially since fish move in 3D space.
    \item \textbf{Turbidity:} Changes in water clarity, due to factors like plankton, can lead to false positives in fish movement tracking.
    \item \textbf{Biofouling:} Algae and dirt buildup on lenses in contact with seawater degrade image quality.
    \item \textbf{Periodic and Multimodal Background:} Algorithms must handle background movements and variations, including periodic movements like those of plants in tides, to avoid misidentifying non-fish objects as targets.
\end{itemize}
Considering the intricate nature of computer vision applications in aquaculture, considerable research efforts have been dedicated to creating algorithms designed to effectively address these specific challenges.

\subsubsection{Biomass Detection of Fish}
In aquaculture, accurately estimating fish biomass is crucial for effective management and sustainable production.  For size estimation, researchers have used algorithms on datasets like ImageNet and the Atlantic fish dataset. \citet{monkman2019using} developed an R-CNN model for estimating European bass length, achieving an average deviation of 2.2\%. However, this model's applicability is limited, especially for overlapping fish. \citet{li2020comparative} used a CNN model based on transfer learning for small sample pond fish data, reaching an accuracy of 93.93\%. \citet{voskakis2021deep}, developed and deployed a CNN model to estimate the length of European seabass and gilthead seabream in cages with relative error of 3.1\% and 7.4\%, respectively 

Accurate weight estimates are crucial for farmers since the weight of fish determines the price received at harvest time, impacting the economic viability of aquaculture operations. Therefore, precise weight estimation is essential for effective planning, resource allocation, and maximising the returns from the harvest.
Typically, researchers focus on predicting quality based on fish body characteristics, using image processing to extract size and shape features. \citet{fernandes2020deep} combined linear regression and CNN for predicting weight from body area, while \cite{saberioon2018automated} used infrared light for image collection and RF and SVM models for back geometry prediction, overcoming lighting constraints. 

\subsubsection{Recognition and Classification of Fish}
Fish farmers need comprehensive information on each fish's attributes such as size, weight, gender, maturity stage, and skin color throughout their growth phases to efficiently track their development and manage stocks \citep{saberioon2017application}. Machine learning facilitates automated fish identification, extracting their features, and classifying them through conventional computer vision methods. For instance, the Stingray system employs computer vision to detect and eliminate individual lice on salmon, addressing parasitic infections in aquaculture \citep{bui2020salmon}.

The task of fish target detection in aquaculture has seen significant advancements with the application of deep learning models like YOLO (You Only Look Once) and CNN (Convolutional Neural Networks). YOLO-based methods have been employed for fish detection in various contexts, such as fish count \citep{armalivia2021automatic} and trajectories \citep{mohamed2020msr}. These methods have been further refined, for instance, by \cite{cai2020modified}, who replaced the Darknet-53 network in YOLOv3 with MobileNet, enhancing feature extraction and overall accuracy. Additionally, CNN models have been increasingly favoured for their efficacy in fish detection. \cite{villon2018deep} used the GoogLeNet architecture combined with soft-max classification, achieving an accuracy rate of 94.9\%. Extensions of CNN models have further improved accuracy, with \cite{rauf2019visual} deepening convolutional layers to reach a maximum accuracy of 96.63\%.

Beyond basic CNN architectures, advanced models like R-CNN and LSTM have also been applied to fish detection. \cite{li2016accelerating} utilised the Faster R-CNN model, initialised with the pre-trained Zeiler and Fergus model, to enhance detection speed and accuracy. Techniques like background subtraction have been employed to distinguish fish from other objects, with varying degrees of success \citep{shevchenko2018fish, gaude2019fish}. Transfer learning methods, as used by \cite{tamou2018transfer}, have achieved high accuracy in challenging underwater environments. Additionally, innovative methods like 360-panoramic systems and binocular vision-based image processing have been developed, further expanding the scope of aquatic species recognition, including live crabs and shrimp, with high accuracy rates reported \citep{cao2020prediction, hu2020shrimp}. These diverse approaches demonstrate the evolving landscape of target detection in aquaculture, offering valuable insights for future research in low-resolution image recognition and underwater species identification.

\subsubsection{Behavioural Analysis}
Accurate recognition of fish feeding behaviour can help farmers optimise feeding, reduce costs, and enhance economic efficiency \citep{lopes2017fish}. \cite{zhou2019evaluation} have made significant strides in this area, employing near-infrared imaging technology to assess fish feeding behaviour. By applying SVM and the gray gradient symbiotic matrix, they quantified the feeding behaviour index, achieving a high correlation (0.945) with expert manual scores 

Parallel efforts by other researchers have also focused on using neural network models for analysing fish feeding behaviour. \cite{maaloy2019spatio} integrated CNN and LSTM to differentiate between feeding and non-feeding behaviours in salmon, achieving 80\% accuracy. \cite{adegboye2020incorporating} further evaluated feeding behavior using Fourier descriptor threshold analysis, based on Noda and Gleiss's prior research dataset, reaching a remarkable 100\% accuracy. \citet{georgopoulou11swimming} developed a continuous, real-time monitoring system to analyze and guide the feeding behaviour for seabass, based on various machine learning models (YOLO and Deep SORT) and computer vision techniques.
These developments underscore the potential of intelligent systems for precise fish feeding management.

Regarding group behaviour in aquatic ecosystems, individual actions based on local stimuli can lead to collective patterns, displaying remarkable coordination and adaptability within groups \citep{giardina2008collective}. Understanding fish group behaviors offers deeper insights into environmental or social changes within these groups \citep{odonncha2021data}. \cite{han2020fish} introduced a group behaviour recognition method using CNN and spatio-temporal information fusion, successfully identifying six distinct group behaviours in fish with an average accuracy of 82.5\%. 

Additionally, the monitoring of abnormal behaviours in fish is crucial for non-invasively assessing fish welfare in aquaculture \citep{papadakis2012computer, georgopoulou2022european}. \cite{zhao2018modified} developed an RNN-based method for detecting abnormal behaviours in densely populated fish schools. In addressing issues like dissolved oxygen anomalies and disease infections, \cite{zhao2016spatial} and \cite{morimoto2018study} proposed machine vision-based assessments for abnormal behaviour in tilapia and shrimp, focusing on indicators such as swimming speed and movement distance. These approaches underscore the value of behavioural analysis in enhancing fish welfare and management in aquaculture settings.

\subsection{AI to Forecast Environmental Conditions}
The increasing integration of AI and digital twin technologies is revolutionising environmental management and decision-making processes. Destination Earth is a standout initiative in this domain \citep{destine_destination_2023}. As a critical part of the European Green Deal, it aims to steer the EU towards a climate-neutral economy by 2050, employing advanced digital models to simulate Earth's systems and support eco-friendly policies \citep{bauer2021digital}.

At the heart of Destination Earth is the Digital Twin of the Ocean \citep{edito_2024}. This project is focused on constructing a high-precision digital replica of the ocean, facilitating precise simulations of oceanic processes. The Digital Twin of the Ocean is multifaceted: it not only offers vital insights into the health of the oceans but also aids in comprehending the effects of climate change on marine ecosystems. Importantly, it assists in the sustainable management of marine resources. By modelling various environmental scenarios, this digital tool enables policymakers to forecast the impacts of different policy choices, ensuring alignment with the European Green Deal's goals, such as minimising ocean pollution, managing fisheries sustainably, and conserving marine biodiversity \citep{tzilivakis_tale_2022}.

Furthermore, the Digital Twin of the Ocean exemplifies the potential of AI and high-performance computing in fostering ecological sustainability and enhancing climate resilience \citep{tzachor2023digital}. Fish farms are particularly significant in the context of the Digital Twin of the Ocean, serving as both a primary source of ocean data and a critical use case for the development and application of ocean technologies. While there has been extensive research into using AI for ocean forecasting and water quality prediction \citep{odonncha2024forecasting}, particularly in freshwater systems \citep{oruche2021transfer, oruche2023attention}. Despite promising research, operational deployments of AI technologies in aquaculture remain limited, underscoring the need for robust infrastructure and standardized practices \citep{heimbach2024crafting}. These studies have concentrated on predicting ocean currents, waves, temperature, dissolved oxygen levels, and other key water quality indicators.

\cite{yang2021computer} summarises some of these applications across univariate and multivariate implementations as well as forecasting different variables

\subsubsection{AI Applications for Environmental Prediction}

The application of machine learning (ML) for time-series forecasting has seen extensive exploration in a broad spectrum of application papers. For classical (linear) time series, where all pertinent information about the next time step is encapsulated in the recent past, a multilayer perceptron (MLP) is typically adequate to model short-term temporal dependence \citep{gers2001lstm}. In contrast to MLPs, convolutional neural networks (CNNs) autonomously identify and extract features from the input time series. Unlike MLPs and CNNs, recurrent neural networks (RNNs) explicitly allow for long-term dependence in the data to persist over time, a feature that is particularly desirable for modelling time series. A comprehensive review of many of these approaches for time series-based forecasting is provided in \citep{esling2012time, gamboa2017deep, fawaz2019deep}.

Owing to the significant computational overhead of physical models, there is an increasing trend towards the application of data-driven DL or ML methods to model physical phenomena \citep{bezenac_deep_2017,wiewel_latent_2018}. Indeed, the geosciences were among the earliest adopters of machine learning techniques, with the widespread use of Kriging techniques \citep{williams2006gaussian}. Kriging, effectively Gaussian Process regression, is a method of interpolation governed by prior covariances and was first introduced by \cite{matheron1963principles} for estimating the most probable distribution of gold based on samples from a few boreholes. It later made significant contributions to the field of geostatistics \citep{dramsch202070}. The interest in the potential of machine learning to address environmental challenges has flourished in the past decade, driven by the growth in data-intensive computing \citep{hey2009fourth}. A central focus has been climate change, and a recent review offers a comprehensive discussion on the potential impact of various machine learning methods for modelling, engineering, and mitigation to address this issue \citep{rolnick2019tackling}.

There is a vast body of literature related to machine learning-based forecasting methods. These include an array of shallow (e.g., decision trees, random forests, support-vector machines) and deep neural network (DNN) approaches. \cite{wolff2020statistical} compared the performance of different statistical, machine learning, and DNN methods for forecasting sea surface temperature (SST). The results indicated that, given sufficient training data volumes, simpler methods could achieve accuracy comparable to DNNs.

Numerous research efforts have developed frameworks to effectively capture the spatiotemporal characteristics of geophysical systems. The Bidirectional Long Short-Term Memory (BiLSTM) networks, with their distinctive architecture, are particularly adept at modeling these characteristics, as they efficiently handle complex dependencies inherent in time series data, encompassing both spatial and temporal dimensions. \cite{ odonncha2022spatio} leveraged BiLSTM networks for predicting temperature and dissolved oxygen levels, utilizing data from a network of IoT sensors. This application underscores the efficacy of BiLSTM networks in processing and interpreting the intricate spatiotemporal patterns present in environmental data.

A widely used framework combines CNN with LSTM to represent both the spatial (CNN) and temporal (LSTM) dependencies within the data. The fundamental structure of this approach is to formulate both the inputs and outputs as spatiotemporal sequences (or a sequence of images). By initially processing the spatial datasets with a convolutional layer, local correlation and global coherence of each pixel are expressed and retained by patches with fixed dimensions, which are then processed by an LSTM that aims to encode the temporal dependencies. This approach has been applied to various geoscientific tasks, such as precipitation nowcasting from rainfall radar maps \citep{Xingjian2015} and forecasting sea surface temperature from satellite-derived observations \citep{yang2017cfcc}.

More recently, \cite{xiao2021dual} used a convolutional LSTM algorithm for multivariate time series prediction. The exogenous features were treated as an "image" and processed with a convolutional layer that extracted information on the relationship between different exogenous features at a point in time. The convolved features were then fed to the LSTM model to learn the temporal dependencies. This framework was applied to time series data from environmental, energy, and financial applications.

A number of studies have investigated data-driven approaches to provide computationally cheaper surrogate models, applied to wave forecasting \citep{james_machine_2018, ODonncha2018}, air pollution \citep{hahnel2020using}, viscoelastic earthquake simulation \citep{devries_enabling_2017}, and water-quality investigation \citep{arandia_surrogate_2018}. Many studies have explored how machine learning could improve ocean monitoring and forecasting either by mining large ocean datasets \citep{Gokaraju_machine_2011} or relating future conditions to historical observations \citep{wolff2020statistical}. These include DL-based approaches to increase the resolution of satellite imagery through down-scaling techniques \citep{Ducournau_deep_2016}; and data-mining applied to the large datasets generated by ocean monitoring and modelling tools to identify pertinent events such as harmful algal blooms \citep{Gokaraju_machine_2011}.

An alternative approach aims to embed information from physics or heuristic knowledge within the network. Physics-informed DL is a novel approach for integrating information from physics. This method approximates the quantity of interest (e.g., governing equation variables) by a deep neural network (DNN) and embeds the physical law to regularise the network. Training the network is thus equivalent to the minimisation of a well-designed loss function that contains the PDE residuals and initial/boundary conditions \citep{rao2020physics}.

\section{AI and Emerging Trends in Aquaculture}
AI applications in aquaculture are at a nascent stage. Technologies such as computer vision, IoT, and deep neural networks offer immense promise. However, operational deployments at scale are still limited. 

Initiatives like Destination Earth \citep{destine_destination_2023} and the Digital Twin of the Ocean \citep{tzachor2023digital} are becoming critical enablers for improved decision-making in ocean-related activities. In parallel, the weather forecasting community has experienced rapid advancements in AI-backed forecasting in recent years. This significant increase in AI implementation, often referred to as a "quiet revolution," has marked a notable milestone where AI has, for the first time, outperformed traditional weather prediction models in terms of efficacy \citep{maskell_rise_2023}. These are certainly indicative of future technology trends related to digital twin solutions for aquaculture\citep{ alvarez2024promoting}

The world’s top weather agency, the European Centre for Medium-Range Weather Forecasts (ECMWF), has embraced the technology: Recently it began to generate its own experimental operational forecasts based on AI methods. Three different architectures are used: 
\begin{itemize}
    \item \textbf{FourCastNetv2-small:} An advanced version of the Nvidia-led FourCastNet deep learning system, which employs Spherical Fourier Neural Operators for spatial dependency analysis. This streamlined model is designed to be compatible with a single Nvidia A100 40GB for inference purposes. Trained on ERA5 data to reduce mean-squared-error in predictions, it functions at a resolution of 0.25° \citep{bonev2023spherical}.
    \item \textbf{Graphcast:} Developed by Google Deepmind, this deep learning system utilizes a graph neural network structure, incorporating an encoder-processor-decoder design with a multi-mesh approach. The model, trained on ERA5 data at a 0.25° resolution, was further refined using the ECMWF HRES forecast to lower mean-squared-error in its predictions \citep{lam2022graphcast}.
    \item \textbf{Pangu-Weather:} Created by Huawei, this deep learning system is based on a 3D transformer architecture to analyse spatial dependencies. It comprises multiple models for making predictions over various time frames, such as 24 hours and 6 hours. The training, conducted with ERA5 data, focused on reducing mean-squared-error for each predictive model, operating at a resolution of 0.25° \citep{bi2023accurate}.
\end{itemize}
These significant advancements can inform algorithms and data pipelines that can enhance digital twin opportunities for aquaculture. Across these models Graph Neural Networks (GNNs) and transformer architectures play key role. 

\subsection{Graph Neural Networks}
Graph Neural Networks (GNNs) represent a category of neural networks designed to work with graph-structured data, adept at handling intricate and interlinked relational data. Their ability to process data represented not as sequences or grids but as networks of interrelations makes them exceptionally suitable for fields where data is network-centric, like environmental and biological systems, or transportation networks. An illustrative case is the GraphCast weather forecasting system, which employs GNNs to effectively handle data with inherent spatial structures.

One of the key strengths of GNNs is their versatility in various scientific and industrial applications. They have been instrumental in drug discovery, where they help in predicting molecular interactions \citep{stokes2020deep}, and in traffic forecasting, where they analyse and predict traffic patterns in transportation networks \citep{derrow2021eta}. GNNs also find applications in social network analysis, where they can be used to model and predict user behaviour and interactions within the network. This broad applicability of GNNs stems from their ability to model complex relational patterns and dependencies, which are common in many real-world datasets.

In the realm of net-pen fish farming, GNNs present significant potential for enhancing various aspects of aquaculture through their advanced capability in modelling complex, interrelated data. GNNs, by their very nature, are adept at capturing relational information, making them particularly suited for representing and analysing the intricate network of interactions within fish farming ecosystems. For instance, in water quality monitoring, GNNs can be utilised to predict parameters like dissolved oxygen, temperature, and nutrient levels. 

Parasitic infections, like sea lice in salmon farming, are naturally amenable to graph-based modelling since the spread of lice between cages, farms, or even regions can be represented as a dynamic network of spatial and biological interactions—making Graph Neural Networks (GNNs) well-suited for capturing such relational structures\citep{langbridge2023causal}. This approach enables predictive analytics for early detection and intervention, potentially reducing the spread of parasites and minimising the need for chemical treatments, thus promoting a more sustainable and environmentally friendly approach to aquaculture.

Moreover, GNNs hold potential in broader industry aspects such as fish behaviour analysis, yield prediction, and feed optimisation. By analysing the spatial interactions and movements of fish within cages, GNNs can provide insights into fish behaviour, stress levels, and overall well-being. Such analyses can inform improved cage management practices, leading to enhanced fish health and growth rates. Further, construction of an effective graph topology informing similarity between cages, farms, or regions can enhance the predictive skill of models \citep{azmat_forecasting_2023, ba2023efficient}. This is particularly critical for aquaculture where factors like ocean currents play a key role in material transport between cages and farms.

\subsection{Large Language Models}
Large Language Models (LLMs), also referred to as foundation models or generative AI, represent a paradigm shift in artificial intelligence due to their scalability, adaptability, and wide applicability across domains. These models, trained on extensive and diverse datasets, are capable of learning a broad range of tasks without task-specific training data. Examples include models like GPT-4 \citep{achiam2023gpt} for natural language processing and early efforts such as BERT \citep{jawahar2019does} for text understanding. Their training relies on vast amounts of data, often scraped from the internet, enabling them to develop a wide-ranging understanding of human language, behaviour, and knowledge. Once trained, LLMs can be fine-tuned or adapted to specific tasks, making them highly versatile and efficient for various applications, from language translation and content generation to more complex tasks like answering questions, writing code, or creating artistic content \citep{bommasani2021opportunities}.

The widespread success of LLMs can be attributed to three main factors \citep{tunstall2022natural}:
\begin{itemize}
    \item \textbf{Transformer Architecture:} Introduced in the 2017 paper "Attention Is All You Need" by Google researchers \citep{vaswani2017attention}, transformers represent a paradigm shift in neural network design. They replace the sequential processing of RNNs/LSTMs with a parallel approach, using an `attention mechanism' to assess the relevance of different parts of input data. This allows for processing entire data sequences simultaneously, enhancing efficiency and effectiveness for tasks with large sequences, like language modelling and translation. Their ability to manage long-range dependencies with reduced computational complexity, while preserving context, has established transformers as the de facto standard for modern natural language tasks. 
    \item \textbf{Pre-trained foundation models:} Instead of building a model from the ground up for specific applications, it is now feasible to use a model pre-trained on a general dataset and fine-tune it for particular needs. This approach reduces the need for large task-specific datasets and extensive training from scratch, which is why these models are referred to as 'foundation models' \citep{bommasani2021opportunities}. They serve as a versatile starting point for crafting task-specific models.
    \item \textbf{Accessibility through Model Hubs like Hugging Face:} The advent of Hugging Face's Transformers library marked a significant development. Earlier, finding pre-trained models was challenging as they were scattered across various platforms. Hugging Face's open-source library supports TensorFlow and PyTorch and simplifies accessing, configuring, fine-tuning, and evaluating state-of-the-art pre-trained models from the Hugging Face Hub. As of 2025, Hugging Face hosts over 900,000 models on its platform, and serves more than 7 million users and 50,000 organisations, including major tech companies like Google and Microsoft \citep{Wolf_huggingface_2025}
\end{itemize}

Transformer-based architectures, initially celebrated for their prowess in language tasks, have expanded their versatility to other sectors including code generation \citep{jimenez2023swe}, computer vision, time series forecasting \citep{liu2025rnns}, and industrial applications. For example, IBM Research has innovated a geospatial model using these architectures that can interpret satellite images to map environmental changes and natural disasters \citep{mukkavilli2023ai}. Utilising these foundation models, developers and researchers can bypass the extensive time and resources traditionally needed to develop new models from the ground up. These models offer a strong foundational skill set that can be customised for specific needs or industries.

While still emerging in aquaculture, the successful application of transformer architectures and foundation models in other industries suggests strong potential for future adoption. Given the superiority of transformers over RNNs, LSTMs, and other deep learning models in tasks like sequential language modeling, it is reasonable to expect their increasing adoption in areas like digital twin technology for fish farming and enhanced automation.

\subsection{LLM Powered Autonomous Agents}
The integration of large language models (LLMs) as autonomous agents is creating transformative opportunities across industries, including aquaculture. By leveraging the natural language capabilities of LLMs, these agents can interpret, process, and respond to complex queries, making them uniquely suited for managing data-rich and dynamic environments. In aquaculture, where operations depend on precise environmental data, resource optimization, and real-time monitoring, LLM-powered agents can serve as both decision-support tools and proactive managers, bridging the gap between data sources and actionable insights.

In an LLM-powered autonomous agent system, the LLM functions as the “brain” of the agent, supported by several essential components \citep{weng_llm_2023}:
\begin{itemize}
    \item Planning: The agent breaks down large, complex tasks into smaller, manageable subgoals, enabling it to handle intricate processes efficiently.

    \item Memory: The agent utilizes \textit{in-context learning}—often facilitated by prompt engineering—to temporarily retain and apply relevant information, functioning as a form of short-term memory. This can be augmented with external vector storage for long-term memory capabilities

    \item External API and Tool Calling: The agent learns to call external APIs for information that is unavailable within its model weights, which are typically static after pre-training. This includes access to current data, the ability to execute code, and retrieval from proprietary databases, expanding the agent’s knowledge and capabilities beyond its pre-trained state.
\end{itemize}

\begin{figure}[t]%
    \centering
    \includegraphics[width=0.99\columnwidth]{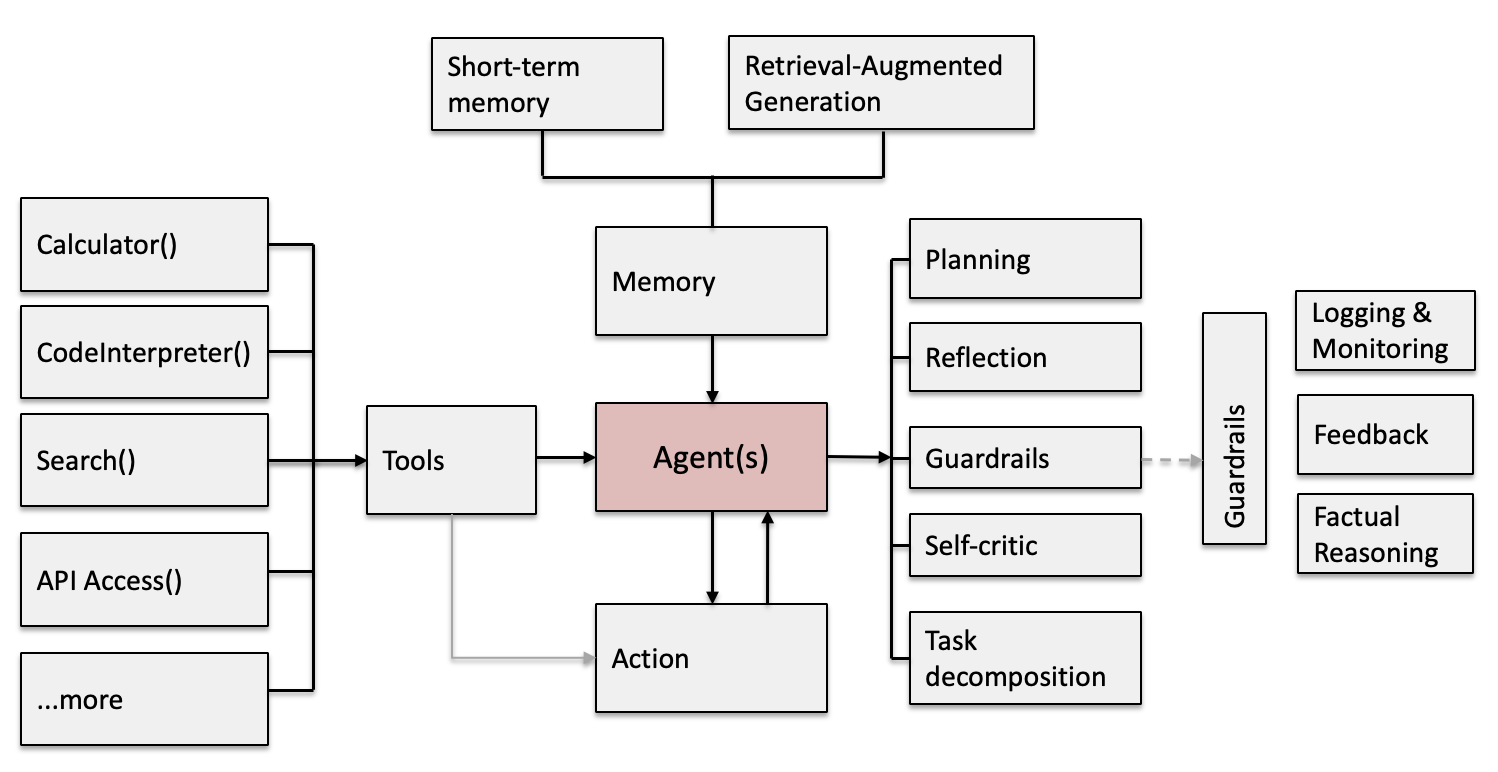}
    \caption{This diagram illustrates the key components and data flows in a large language model (LLM)-driven agent architecture. The central agent coordinates interactions between tools (e.g., calculators, code interpreters, APIs), memory systems (short-term and retrieval-augmented), and action modules. Cognitive processes such as planning, task decomposition, self-critique, and reflection support goal-oriented behavior and adaptive reasoning. Guardrails ensure system robustness through safety, factual reasoning, feedback, and monitoring. The architecture enables scalable, flexible execution of complex tasks in dynamic environments. }%
    \label{fig:autonomouse agent}
\end{figure}

These components work together to enhance the LLM's performance, enabling it to plan complex tasks, remember information effectively, and extend its functionality through external tools. Figure \ref{fig:autonomouse agent} summarises these capabilities.

One of the most significant advantages of LLM-powered agents is their ability to perform "tool calling" —accessing and interacting with various tools and systems based on the needs of a particular task. In aquaculture, this functionality allows an agent to autonomously query databases, access web-based information, interact with APIs, and analyze incoming data streams. By invoking specific tools, such as:
\begin{itemize}
    \item Databases: Agents can access internal databases to retrieve historical information on fish health, water quality, and feeding patterns, enabling real-time analysis and historical comparisons.
    \item Web Search: With access to external web sources, agents can find recent research, industry best practices, and updates on regional regulations or environmental concerns.
    \item Weather Data: By accessing weather forecasts and live updates, LLM agents can offer valuable insights into weather conditions, allowing managers to anticipate challenges related to water temperature, storms, and seasonal changes that may impact fish welfare.
    \item Environmental and Model Data: Agents can interface with environmental sensors and model-based systems, providing real-time updates on factors like dissolved oxygen levels, salinity, and currents, all of which are critical for maintaining optimal conditions within aquaculture systems.
\end{itemize}
The ability to interact with these varied data sources positions LLM agents as versatile tools capable of synthesizing information from multiple channels to create a cohesive and context-aware picture of operations. This capability is particularly valuable in complex systems like aquaculture, where diverse and interdependent variables influence outcomes.

LLMs' fluency in natural language enables their use as a universal interface, making them accessible to a wider range of users, including those without technical training. This aspect greatly enhances accessibility and usability, as non-technical personnel in aquaculture can interact with the system simply by using conversational language. For example, an aquaculture manager could ask the LLM-powered agent, “What’s the expected water temperature over the next week?” or “Have there been any significant changes in dissolved oxygen levels in the last 24 hours?” The agent can interpret these questions, identify relevant data sources, retrieve the necessary information, and provide a straightforward response.

This language-based interface also facilitates more complex interactions, such as requesting a summary of recent environmental conditions and their impact on fish growth rates or asking the agent to compile a report on feeding efficiency based on recent patterns. The agent’s ability to understand and execute these natural language commands allows for rapid decision-making and reduces the learning curve for users, promoting broader adoption of technology in aquaculture operations.

The use of LLM-powered agents in aquaculture remains nascent, but their rapid development suggests significant potential for expanding their capabilities across monitoring, prediction, and autonomous decision-making. Key areas of potential growth include enhanced integration with remote sensing technologies, refined models for predicting fish growth and health outcomes based on environmental factors, and deeper incorporation of machine learning algorithms to enable predictive analytics. As LLM agents continue to evolve, their ability to learn and adapt to specific aquaculture contexts could make them invaluable as autonomous managers capable of optimizing operations with minimal human oversight.

Overall, LLM-powered autonomous agents offer a compelling vision for the future of aquaculture, providing a centralised, language-driven interface that leverages diverse data streams and advanced tool integrations. By harnessing these capabilities, aquaculture operations can become more resilient, adaptive, and efficient, paving the way for sustainable growth in the face of environmental challenges and market demands.

\section{Case Studies}

While AI and machine learning applications in aquaculture are rapidly expanding, much of the current work remains exploratory or limited to isolated deployments. Unlocking the full value of these technologies depends on building robust data infrastructures, scaling models effectively, and integrating AI seamlessly into existing operational workflows. Advances in areas such as unmanned monitoring platforms, remote sensing, edge computing, and simulation offer exciting possibilities for accelerating the adoption of AI-driven solutions. Notably, the convergence of machine learning with digital twins and real-time environmental modelling stands to transform aquaculture from reactive management to proactive, predictive control.

The following case studies illustrate how some of these emerging technologies are already being applied in practice. They highlight different strategies to improve environmental forecasting and operational decision-making in marine and aquaculture settings. Specifically, the case studies demonstrate:
\begin{itemize}
    \item The deployment of real-time sensor networks and machine learning models to forecast dissolved oxygen on Atlantic salmon farms in Scotland;
    \item The use of satellite remote sensing combined with machine learning for spatial estimation of water quality variables in aquaculture zones;
    \item A novel LLM-based agentic architecture applied to digital twin workflows for shipping, showcasing how agentic AI can support complex operational decision-making.
\end{itemize}
Together, these examples provide a snapshot of how AI and digital twin technologies are beginning to reshape the aquaculture and marine sectors.

\subsection{Case Study 1: Application of a real time sensor network in developing machine learning models to forecast dissolved oxygen on Atlantic salmon farms in Scotland}

\subsubsection*{Introduction}
Oxygen is a critical and dynamic parameter that affects fish metabolism and stress and ultimately impacts growth efficiency \citep{remen2012effects}. Widescale environmental monitoring on aquaculture farms has been adopted by the industry to monitor the complex dynamics of dissolved oxygen at the pen level and is used to inform feeding decisions \citep{burke2021oceanographic}. This adoption has created long term environmental data sets that can be used in model development to forecast oxygen in open net pens. Oxygen forecasting can help to improve  decision-making to maximize growth efficiency. During feeding, oxygen levels must be above a given threshold (which varies depending on temperature but generally, 70\% saturation is used) and there must be reasonable confidence that oxygen will remain above this threshold hours later, during the digestive process that uses up additional oxygen \citep{remen2016oxygen}. If oxygen levels are too low, fish become stressed and energy is allocated to other more critical life functions instead of growth, and in severe cases fish will regurgitate feed resulting in feed waste and reduced feed conversion ratios (FCRs). Providing confidence in near-term oxygen levels using machine learning predictions can assist with real time decision making and helps to maximize growth and enhance welfare by feeding only under optimal environmental conditions.

Machine learning models have been proposed as an alternative to traditional ocean models to capture the fine-scale processes that a pen-level forecast necessitates and to account for the effect of the biomass on dissolved oxygen levels in the pen. This project scrutinizes which machine learning models best capture oxygen dynamics at the pen level and examine the robustness of a model as it is scaled up and tested on different pens within the same farm, on pens at different farms within the same geographic region and on pens at farms in different regions. The model was first trained on a farm located in the Shetland Islands, and then tested on a farm on the west coast of mainland Scotland and finally performance was evaluated on a farm located on the west coast of Canada. This work is ongoing, but some preliminary results are presented below.

\subsubsection*{Methodology}
Preprocessing of operational data is a critical step as datasets often contain many imperfections such as erroneous measurements, anomalies, and missing data. The data consisted of historical observations of oxygen concentration, temperature, and salinity for individual cages on various farms. During preliminary analysis, temperature and salinity were identified as key features for normalization and time-based variables such as the day of the week, hour of the day, and month of the year were also incorporated \citep{linardatos2020explainable}. These variables are crucial as they enable the model to effectively capture the trends and seasonality of oxygen levels. A circular numeric encoding strategy was adopted for time-based variables to maintain their cyclical nature, enhancing the model's ability to discern patterns in oxygen levels. Many different models were tested and compared including linear regression, random forest, and Long Short-Term Memory (LSTM). However, the Temporal Fusion Transformer (TFT) model reported best performance due to its effectiveness in handling complex time-series data. The adoption of this architecture was informed by several standout features:  
\begin{itemize}
    \item Multi-headed Self-attention Mechanism: This feature allows the model to assign differential significance to various temporal intervals, a critical function for decoding complex time-related patterns in data, such as seasonal fluctuations and the influence of specific events on oxygen concentration. By prioritizing relevant periods over others, the model achieves superior forecasting precision. 
    \item Dual-stage Architecture: The model employs an encoder-decoder framework that combines historical data analysis with future event prediction. The encoding phase processes past data, whereas the decoding phase is dedicated to forecasting the target time series. This method increases the model's understanding of time patterns, improving its prediction accuracy and reliability for different forecasting horizons. 
    \item Quantile Forecasting: This enables the prediction of a spectrum of outcomes at varying probability thresholds. By forecasting seven quantiles of future oxygen concentrations, the model provides a comprehensive overview of potential scenarios, facilitating enhanced decision-making capabilities. Seven is the standard number of quantiles for this model and uses an uneven distribution of thresholds for the output (0.02, 0.1, 0.25, 0.5, 0.75, 0.9, 0.98).
\end{itemize}

\begin{figure}[t]%
    \centering
    \includegraphics[width=0.99\columnwidth]{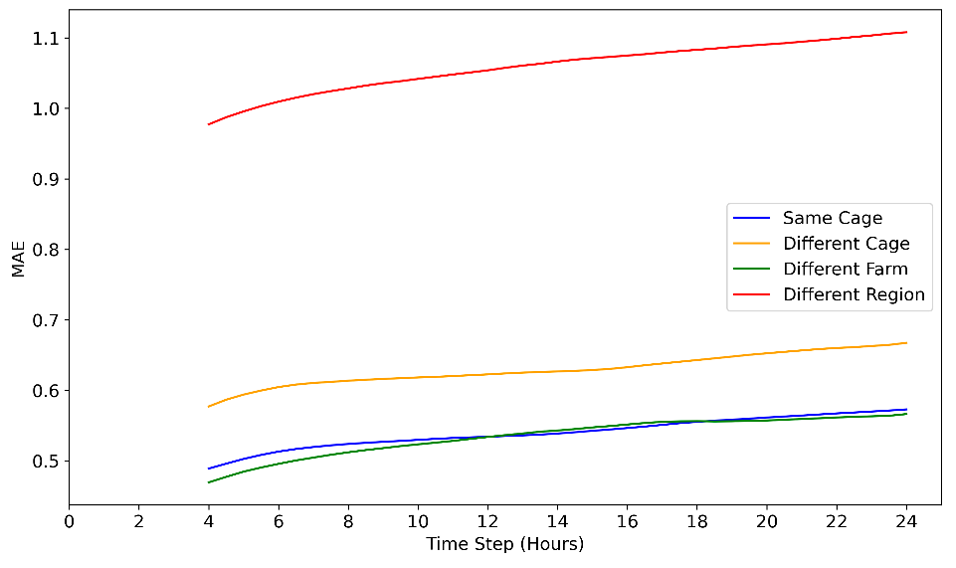}
    \caption{Mean average error (MAE) of 24-hour prediction for various tests of the model on the same cage it was trained on, a different cage from the same farm, a cage from a different farm within the same region and a cage on a farm in a different region. }%
    \label{fig:mae24}
\end{figure}

The refinement of the model involved a detailed evaluation of its forecasting performance, at the individual pen level and across entire farms, through comparative analysis with baseline models, utilizing its dual-stage and multi-horizon forecasting functionalities. The models were trained using a structured cross-validation method to verify its efficiency across assorted data segments, including different cages, farms, and geographical regions. This approach, coupled with strategic parameter selection, demonstrated the model's adaptability and applicability across different farms. To assess the model's predictive performance, particularly its ability to forecast extreme changes in oxygen levels, a set of evaluation metrics was carefully chosen. These included mean absolute error (MAE) and quantile-based metrics to evaluate accuracy within specific ranges (Figure \ref{fig:mae24}).

\begin{figure}[t]%
    \centering
    \includegraphics[width=0.99\columnwidth]{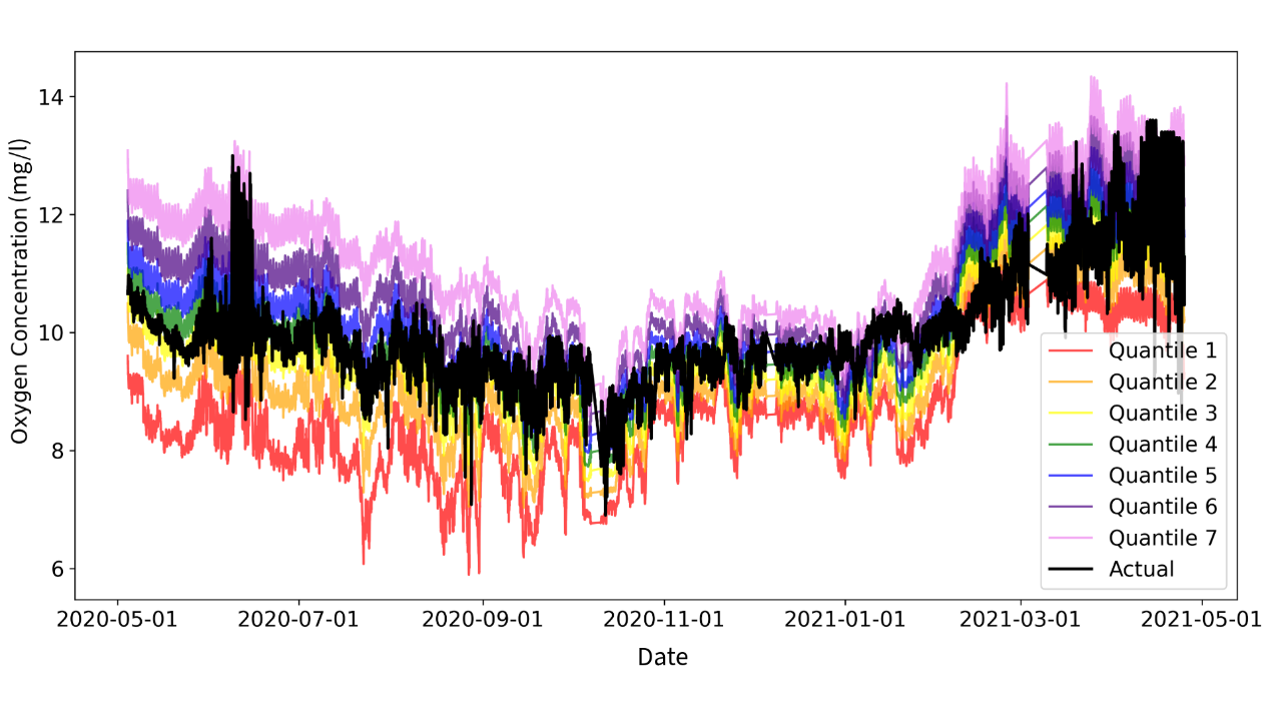}
    \caption{Actual raw oxygen concentration and model output of a 24-hour prediction using 7 quantiles that indicate a confidence interval for the forecasted values. Data presented shows the output of the model on a single cage that was trained on the entire farm dataset.}%
    \label{fig:raw_data}
\end{figure}

Regarding the quantile confidence interval accuracy, the model analyses data at time t and compares it with the quantiles for time t + 24 hours (ranging from the lowest quantile, Q1, to the highest, Q7 ), measuring how often the actual oxygen levels fall within these predicted ranges, providing a clear percentage of coverage (Figure \ref{fig:raw_data}) and allowing operators to derive a range of model outputs based on their risk tolerance. The practical deployment of the model will highlight its utility in generating real-time forecasts. By analysing data from the preceding 48 hours, it can project oxygen levels for the upcoming 24-hour period across seven quantiles, delivering insights into potential variances.

\subsubsection*{Conclusions}
The model was capable of outputting 24-hour predictions for various farms within the same region, with an MAE less than 0.7 mg/l. These findings demonstrate that reasonably accurate pen-level forecasting is achievable, validating ML as a viable alternative to traditional environmental models. The MAE increased when the model was tested on farms in different geographic regions, which suggests that models should be trained on data within a reasonable geographic area to achieve the highest possible accuracy. Operational forecasting is a growing area of research in precision fish farming, aimed at equipping farmers with data-driven tools for more proactive decision-making. Oxygen concentration is a dynamic and complex measurement that is impacted by many distinct factors including, tidal and water exchange, biomass, farm orientation, and plankton presence and this research shows that machine learning models can accurately capture these complexities and there is potential for these insights to be used on farms to help with feeding decisions under challenging conditions. 

This study highlights how machine learning can significantly enhance pen aquaculture forecasting by overcoming the limitations of traditional numerical models, which are bound by their grid resolution and sub-grid process parametrization. Unlike these conventional methods, machine learning models adeptly adjust to the scale of observational data. Coupled with the influx of high-resolution data from Internet of Things (IoT) devices, machine learning enables the development of more detailed and adaptable forecasting frameworks, paving the way for improvements in aquaculture efficiency and sustainability. Further details on this work can be found in \cite{khan2024applying} 

\subsection{Case Study 2: Dissolved oxygen estimation in aquaculture sites using remote sensing and machine learning}

\subsection*{Introduction}
Dissolved Oxygen (DO), together with temperature, is one of the most critical environmental parameters affecting the physiology and performance of marine organisms. Consequently, accurate monitoring is of great importance for finfish aquaculture as it directly affects the survival, growth, food intake, and health of the farmed fish \citep{oldham2019metabolic, burt2012environmental}. For net pen aquaculture in particular, the continuous monitoring of DO is a necessity as environmental conditions at the rearing sites are outside human control and are therefore prone to fluctuations that may threaten production.

DO monitoring has utilised manual, semi-automated, and fully automated in-situ sensors \citep{parra2018physical}. While such modes of DO monitoring are simple and have been effective in the past, the rapid expansion of aquaculture towards precision farming methods necessitates advanced systems that contribute towards real-time monitoring and forecasting \citep{odonncha2019precision, raju2017knowledge}. Accurate forecasting of dissolved oxygen depends heavily on capturing the unique spatial and environmental characteristics of each aquaculture site. While in situ sensors are promising, they do not provide complete coverage at the pen-, farm-, and bay-scale level. Furthermore, organised zones of aquaculture activity with several farms installed have similar needs and large scale monitoring is essential for forecasting. Remote sensing is a powerful tool to enhance farm monitoring and decision making.

Although remote sensing is widely used to detect variables that affect the optical properties of water at specific wavelengths such as chlorophyll-a (Chl-a), total suspended matter (TSS), and temperature, it is unlikely to record the concentration of DO directly from the reflection values of the satellite sensors. Instead, it can be estimated indirectly because of its correlation with other parameters, such as temperature and Chl-a \citep{guo2021generalized, kim2020application}.

DO generally exhibits a negative correlation with sea temperature \citep{matear2003long, mavropoulou2020dissolved}. Primary production also affects DO levels, which is often measured by Chl-a, an indicator of phytoplankton abundance in surface waters \citep{guy1993photosynthetic, lewis2016regional}.

Based on the above theoretical mechanisms, several researchers have tried to develop models for the correlation of DO with environmental parameters, such as sea surface temperature (SST) and Chl-a \citep{karakaya2011monitoring, batur2018assessment, kim2020application, guo2021generalized}. This aims to provide the potential to monitor DO from satellite sensors. Different approaches such as Response Surface Regression (RSR), Multiple Linear Regression (MLR), Artificial Neural Network (ANN), Support Vector Machines (SVM) and Support Vector Regression (SVR) have been scrutinized. While previous research has been successful in estimating DO levels, data sparsity and the demand for high resolution estimation in coastal zones has limited the value of the approach for aquaculture. In our study, we developed a methodology for the estimation of DO in coastal areas with aquaculture facilities, based on daily CMEMS data and machine learning techniques.

\subsection*{Methods and Data}
The satellite data used were from Copernicus Marine Environment Monitoring Service (CMEMS) that provides regular and systematic biogeochemical and physical information on the marine environment. In our study, we exploited Level-4, daily, gap-free satellite observations for Chl-a concentration and SST from multi-platform observations. Our in-situ dataset includes field measures of DO and SST from a fish farm in Lesvos Island, Greece. Dissolved oxygen was measured daily between 8:00am and 10:00am using an OxyGuard sensor with a scavenging packet. 80\% of the dataset served as the training sample, while the remaining 20\% was utilized as the test sample.

The farm is located at Lesvos Island, in North Aegean Sea, an area which is generally characterized as an oligotrophic sea \citep{krom1991phosphorus}. The capacity of the farm is 380T per year and the main fish farmed are the European sea bass (\textit{Dicentrarchus labrax}) and the gilthead sea bream (\textit{Sparus aurata}). Water temperature in the surrounding area ranges between 14°C to 24°C and the salinity is constant at 39 psu. The average depth in the farm is approximately 50m.

For the study, we developed an approach for DO estimation based on a Support Vector Regression machine learning model, using CMEMS data and in-situ observations. For each spatial coordinate, an input array was constructed incorporating chlorophyll-a (Chl-a) and sea surface temperature (SST) values across a three-day lagged window prior to sampling.

Support Vector Regression (SVR) has been demonstrated to effectively tackle both linear and nonlinear problems, showing strong performance even with small datasets \citep{guo2021generalized}. SVR, a variant of Support Vector Machines, is a non-linear, non-parametric machine learning model that adheres to Vapnik's principle of structural risk minimization for supervised learning \citep{vapnik2013nature}. Unlike linear regression, which seeks to minimize squared error sums, SVR aims to manage errors within a predefined tolerance, enabling the selection of error margins ($\epsilon$) and the determination of the optimal hyperplane for decision-making. This approach provides flexibility in model accuracy by adjusting ($\epsilon$) and incorporating slack variables ($\xi$) to tolerate training errors, enhancing the model's resilience to noise and outliers. The model's formulation involves an objective function and constraints, detailed in equations \ref{eq1} and \ref{eq2}, which guide the optimization process:

\begin{equation}
\min \frac{1}{2} \|w\|^2 + C \sum_{i=1}^n |\xi_i|
\label{eq1}
\end{equation}

\begin{equation}
|y_i - w_i x_i| \leq \epsilon + |\xi_i|
\label{eq2}
\end{equation}

where $y_i$ is the target, $w_i$ is the coefficient, and $x_i$ is the predictor (feature). $C$ corresponds to a user-selected parameter to control the complexity of the model, acting as a trade-off parameter between nonlinearity and number of training errors.

\begin{figure}[h]
\centering
\includegraphics[width=0.8\textwidth]{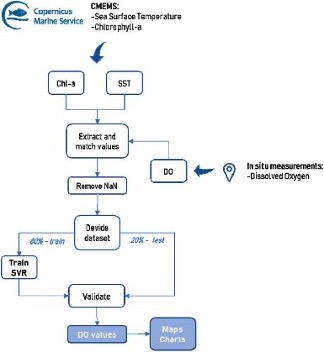}
\caption{Flowchart of the implemented methodology. Reproduced from \cite{chatziantoniou2022dissolved}, \url{https://doi.org/10.1016/j.rsase.2022.100865}, under the CC BY-NC-ND 4.0 licence.}
\label{fig1}
\end{figure}

To capture more intricate hyperplane shapes beyond linear approaches, one can utilize kernel functions like polynomial, radial basis function (RBF), and sigmoid kernels. In the current study, the RBF kernel was chosen for its promising capabilities \citep{guo2021generalized}. SVR, with its simple structure, is supported by various programming languages. Model development was performed in Python 3.4 using the scikit-learn library (v0.18.1), which provides efficient support for SVR and related regression techniques. Model evaluation was conducted using residual analysis alongside mean absolute error (MAE) and root mean square error (RMSE) as standard performance metrics. MAE calculates the precision expressing the average distance between the estimated and the real values, and is described by equation \ref{eq3}. RMSE describes the standard deviation of the residuals (equation \ref{eq4}).

\begin{equation}
\text{MAE} = \frac{\sum_{i=1}^n |y_i - x_i|}{n}
\label{eq3}
\end{equation}

\begin{equation}
\text{RMSE} = \sqrt{\frac{\sum_{i=1}^n (y_i - x_i)^2}{n}}
\label{eq4}
\end{equation}

where $y_i$ is the predicted values and $x_i$ the measured values. The initial dataset included dates, in-situ DO measurements, latitude, and longitude together with the values of Chl-a and SST from CMEMS for each date and the previous 3 days. In Fig. \ref{fig1}, a graphical representation of the approach is presented.

\subsection*{Results}

\begin{figure}[h]
\centering
\begin{minipage}{0.32\textwidth}
    \centering
    \includegraphics[width=\textwidth]{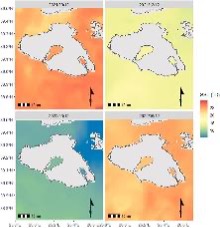}
    \caption*{(a)}
\end{minipage}
\begin{minipage}{0.32\textwidth}
    \centering
    \includegraphics[width=\textwidth]{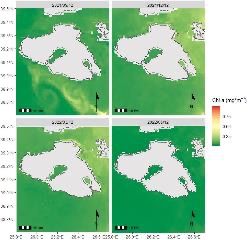}
    \caption*{(b)}
\end{minipage}
\begin{minipage}{0.32\textwidth}
    \centering
    \includegraphics[width=\textwidth]{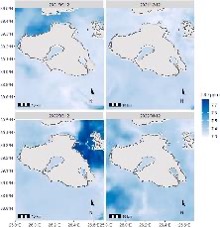}
    \caption*{(c)}
\end{minipage}
\caption{The spatial distribution of (a) SST (b) Chl-a and (c) DO around Lesvos Island for four representative dates (fall, winter, spring, summer). Reproduced from \cite{chatziantoniou2022dissolved}, \url{https://doi.org/10.1016/j.rsase.2022.100865}, under the CC BY-NC-ND 4.0 licence.}
\label{fig2}
\end{figure}

The data collected from CMEMS (Chl-a, SST) and the estimated DO were used to create daily maps for examining their spatial distribution with a resolution of 1 km. Fig. \ref{fig2} presents snapshots of SST, Chl-a and DO in the surrounding sea area of Lesvos.

Fig. \ref{fig2}(c) presents modelled estimates of DO. Naturally, the spatial distribution follows similar patterns to those of SST and Chl-a. Higher concentrations are observed in the north-eastern waters, near the coastline, presenting a clear negative correlation with SST and a weaker positive correlation with Chl-a. In general, DO concentration varied from 7 to 7.8 ppm across the year with four pronounced drops below 6.5 ppm. Highest values were detected in the winter and spring months with lowest values reported during summer and fall.

The accuracy of our model was tested using the value of Mean Absolute Error (MAE) and Root Mean Squared Error (RMSE), as well as plots of the residuals. These are the most widely-used metrics for evaluating a regression model. Both metrics quantify the distance between the actual and the predicted values and are measured in the same units as the output variable, which contributes towards an interpretable performance metric. Our model performed well with MAE of 0.11 and RMSE of 0.13, and the residuals were evenly distributed around 0. These indicate accurate model performance with no obvious bias present.

We have chosen to use MAE and RMSE indices to validate our model as they give a clearer view of the error and distribution of the residuals. On the contrary, $R^2$ was poor (0.32) compared to other studies. $R^2$ is generally used in similar studies to examine the fitness of the model instead of the performance, and it doesn’t adequately consider the magnitude and direction of error. In this context, we do not consider the poor value of $R^2$ to be informative for the practical applicability of the method.

Our results show a promising approach to estimate DO at aquaculture sites, which paves the way for the development of real-time monitoring systems for aquaculture. The methodology successfully correlates DO with SST and Chl-a, reporting high agreement with field data. The model replicates not only shifts in the seasonal patterns of DO but also sudden drops in concentration in at least four occasions during the summer. While the drops in the analyzed time-series are of minor practical interest because they are relatively small and thus, don't constitute a concern for fish, the capability of the method to realistically capture them is reflective of its potential as a monitoring tool that can accurately estimate high resolution dynamics.

Another advantage of the proposed approach is that it increases the spatial scale at which monitoring can occur. While in-situ sensors are paramount for precise measurements within a single cage of a farm, a remote sensing approach offers a better representation of the oxygen conditions, and thus, potential dangers, at a larger area. Considering that aquaculture farms are often aggregated within the limited coastal space of suitable farming areas, either as single operations or as an organized industrial park, it is imperative for the aquaculture producers to monitor the DO at a higher spatial scale. The same applies to administrative authorities that may be interested in the supervision of zones of organized aquaculture activity at a regional level.

\subsection{Case Study 3: Agentic AI for Digital Twin Applications in Shipping}

\subsection*{Introduction}
The global shipping industry, comprising over 100,000 vessels facilitates the transport of 90\% of the world’s goods, yet it also contributes 3\% of global greenhouse gas emissions, which have risen by 20\% in the past decade \citep{unctad2023review, stuchtey2023ocean}. To meet its environmental targets, reducing emissions by 20\% by 2030 and 70\% by 2040, the International Maritime Organization (IMO) prioritizes improvements in ship efficiency \citep{mepc20232023}.

However, these ambitions are complicated by an ageing global fleet, where over half of the vessels are older than 15 years, making them difficult to retrofit yet premature for scrapping \citep{unctad2023review}. In the midst of these constraints, digitalisation and AI offer transformative potential in areas such as route optimization, fuel efficiency, and predictive maintenance \citep{katsoulakos2024shipping}. While platforms like Maersk's TradeLens and Rolls-Royce's autonomous ship initiatives have made strides, the industry's complexity and regulatory hurdles demand more scalable, integrated solutions accessible to a broad range of industry persona \citep{jovanovic2022managing, marine2016autonomous}.

Key priorities include, seamless data integration across fragmented systems, real-time monitoring for predictive maintenance, scalability for diverse fleets, and transparent adherence to regulatory compliance across global frameworks \citep{katsoulakos2024shipping}. Additionally, solutions must provide automation and configurability to address the varied needs of industry personas without relying on specialised expertise \citep{odonncha2024towards}. 

\subsection*{AI Agents}
Artificial Intelligence (AI) agents are systems designed to autonomously perform tasks by perceiving their environment, making decisions, and executing actions to achieve specific goals. Recent advances in large language models (LLMs) have significantly enhanced the capabilities of AI agents, enabling them to handle more complex and dynamic tasks.

LLM-based AI agents can now plan and act by analysing tasks, formulating strategies, and adjusting their actions based on feedback. This mirrors human problem-solving processes, allowing agents to refine their approaches through self-critique and iterative improvements. For example, these agents can autonomously search the web, solve mathematical problems, and even review and correct their work, thereby increasing their effectiveness in various applications.

To manage these complex tasks, engineers are enhancing LLMs with additional modules that improve memory, planning, and tool-utilisation capabilities. This development marks a shift from traditional, monolithic AI models to more sophisticated, multicomponent systems. Such systems integrate various functionalities, enabling AI agents to perform a broader range of tasks with greater autonomy and adaptability.

\section*{Digital Twin for Green Shipping AI Agents}
This case study introduces a digital twin for green shipping (DT4GS) framework that uses large language model (LLM) agents to enhance automation and efficiency in the shipping industry \citep{timms2025agentic}. At its core, the system includes a central reasoning agent that breaks down complex tasks into smaller, manageable steps and coordinates with specialised tools to carry them out. By using natural language as the main interface, this approach allows different users—such as ship owners, captains, engineers, and technicians—to interact with the digital twin in a simple and intuitive way. Each user can query and explore ship systems based on their specific needs, without requiring technical expertise in AI or system modeling.

 We developed a suite of tools specifically designed for industrial applications, enhancing agentic decision-making by providing a comprehensive view of real-time ship operations along with contextual insights. These tools integrate structured information from a domain-specific knowledge graph, real-time sensor data, and technical specifications, including potential failure data. By leveraging LangGraph\footnote{\url{https://langchain-ai.github.io/langgraph/}}, the framework’s modular and adaptable design allows it to meet diverse operational requirements and significantly improve decision-making efficiency. 
 LangGraph is a multi-agent orchestration framework built on LangChain that we adopted to manage tool-using LLM agents in DT4GS. A planner in LangGraph (or in similar LLM-based systems) is a critical component that is responsible for task decomposition, decision-making, and execution sequencing. In AI agent frameworks, planners allow agents to break down high-level objectives into manageable subtasks, prioritise them, and decide in what order and manner those subtasks should be executed. 
 
\begin{figure}[t!]
    \centering
    \includegraphics[ width=\linewidth]{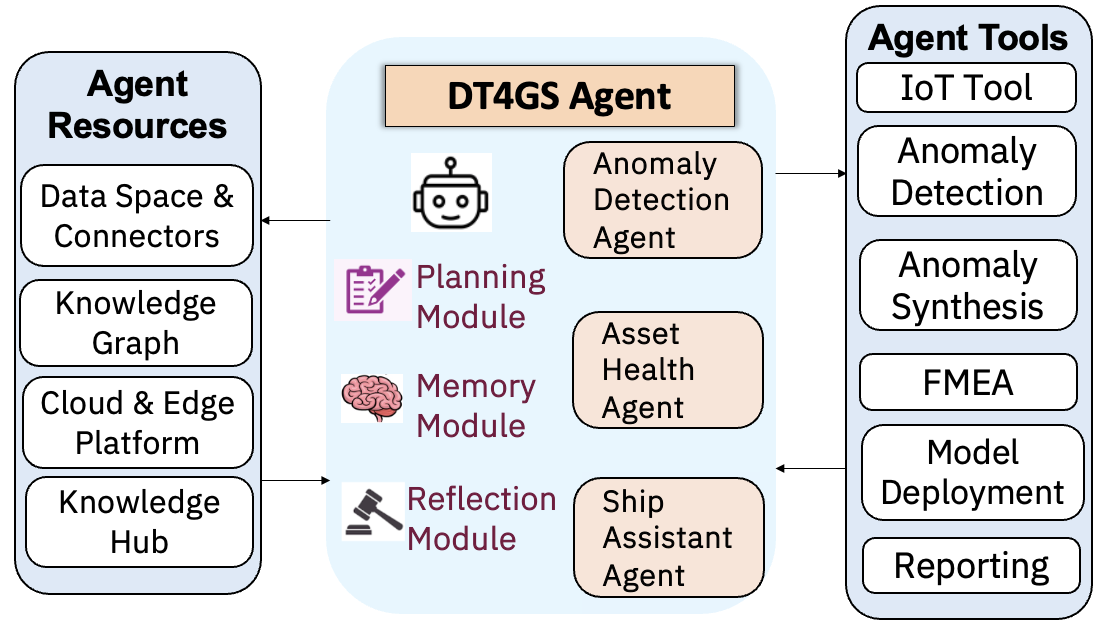}
    \caption{The DT4GS Agent acts as a central orchestrator that coordinates specialized sub-agents and tools to support key operational tasks. It includes a planning module that decomposes tasks and delegates them to appropriate agents (e.g., anomaly detection, asset health, ship assistant). The memory and reflection modules track context and assess system performance. On the left, agent resources such as a data space, knowledge graph, and edge/cloud platforms provide contextual information. On the right, agent tools perform domain-specific functions such as anomaly detection, FMEA, and reporting. This modular, language-driven framework enables flexible, autonomous interaction with complex ship systems.}
    \label{fig:diagram}
\end{figure}

 We demonstrate the capabilities of this framework on real-world shipping use cases, showcasing its practical applicability and scalability.
 Figure \ref{fig:diagram} presents the pipeline that integrates agentic reasoning with a domain-specific knowledge graph and contextual data to streamline the selection, refinement, and execution of digital twin tasks. This framework allows the agent to leverage core digital twin functionalities, including:
 \begin{itemize} 
      \item \textbf{Dataspace}: A centralised repository encompassing sensor and operational data, providing a comprehensive view of various ship operations. 
     \item \textbf{Knowledge Graph}: A structured representation capturing the relationships between the vessel, its components, and associated operational variables. The graph interconnects critical elements such as the main engine, auxiliary engine, and cargo system with key variables like temperature, pressure, and performance metrics, forming a network that reflects the vessel’s functional state. 
     \item \textbf{Model Builder}: A system designed to handle the selection, construction, and refinement of machine learning models tailored to specific task requirements.  It ensures that predictive and prescriptive analytics align with the operational goals of the digital twin.
     \item \textbf{Edge Orchestration Engine}: A component utilising the Linux Foundation's Open Horizon\footnote{\url{https://lfedge.org/projects/open-horizon/}} to efficently manage the deployment and placement of containerised applications across a distributed fleet of edge nodes. This enables real-time processing and decision-making close to the data source.
 \end{itemize}

Agents are equipped with a suite of external tools that extend their ability to interact with digital twin resources and enhance their ability to observe, analyse, and act upon system states. These tools include:

\begin{itemize}
    \item \textbf{IoT Tool}: Facilitates integration with IoT-enabled systems, allowing the agent to retrieve and process real-time sensor data. 
    \item \textbf{Knowledge Graph Tool}: Grants agent access to a domain-specific knowledge graph, enabling complex queries on ship operations, asset relationships, and historical data for enriched decision support. 
    \item \textbf{Data Quality Tool}: Conducts initial data quality assessments by evaluating expected behaviours, identifying inconsistencies, and ensuring data integrity before further processing. Within an agentic workflow, it can generate metrics on data quality, providing users with insights for further refinement and iteration.
    \item \textbf{Anomaly Detection Tool}: An online unsupervised anomaly detection approach that leverages an efficient formulation of the optimal transport (OT) problem in one dimension to detect anomalies in noisy seasonal time series data (described further in \cite{langbridge2024optimal}. 
    \item \textbf{Failure Modes and Effects Analysis (FMEA) Tool}: FMEA is a systematic method that is used to identify and assess possible failure modes within a system, process, or product. It focuses on understanding the causes and effects of each failure mode and evaluating their potential impact \citep{stamatis2003failure}. FMEA helps prioritise risks based on factors such as severity, occurrence, and detectability, enabling preventive measures to be taken. Typical information included in an FMEA consists of the component or system location, potential failure modes, causes of failure, severity of the failure, and the resulting effects of the failure on the system. In industrial sectors like shipping, Failure Mode and Effects Analysis (FMEA) is often mandated by classification societies, including the American Bureau of Shipping \citep{abs_FMEA_2015}. 
    \item \textbf{Time series forecasting tool}: Configures a time series forecasting model to analyse sensor data identified by the agent. This enables predictive assessments of component health based on historical data and relevant contributory features. The model can be deployed in zero-shot mode for immediate inference or further fine-tuned using historical data for improved accuracy.
    \item \textbf{Edge Orchestration Tool}: Automates the generation of JSON policy documents to manage the deployment of applications at the edge on a specific ship. Deployment decisions are based on parameters such as ship name, geographic location, company affiliation, or integration with an IoT tool, ensuring adaptive and policy-driven orchestration of edge computing resources.
    \item \textbf{Work Order Generation Tool}: Automates failure response planning by evaluating identified anomalies and leveraging FMEA insights to suggest potential root causes and recommend mitigation actions. This tool streamlines maintenance workflows and enhances predictive maintenance strategies.
\end{itemize}

\subsection*{Deployment of the DT4GS Agent}

Shipboard asset management and maintenance require a structured workflow that integrates data-driven diagnostics with AI-powered automation. This process involves data retrieval, model inference, results visualisation, and actionable diagnosis. Agentic workflows facilitate an end-to-end solution, reducing manual intervention while enhancing predictive capabilities for shipboard systems.

\begin{figure}[h!]
    \centering
    \includegraphics[ width=\linewidth]{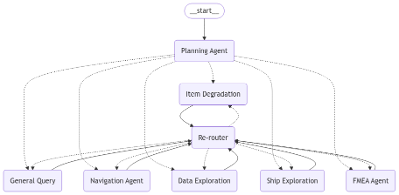}
    \caption{Schematic of the agentic workflow using a planner–router architecture. The Planning Agent parses the initial user prompt and assigns subtasks to specialized agents (e.g., Navigation, FMEA, Ship Exploration). The Re-router coordinates execution flow by determining whether further steps are needed, enabling iterative multi-agent collaboration.
}
    \label{fig:langgraphschematic}
\end{figure}

Figure~\ref{fig:langgraphschematic} illustrates a modular agentic workflow for handling use cases such as these. The process begins with a user prompt that the Planning Agent receives. This agent decomposes the prompt into a series of subtasks and sends them to domain-specialized agents based on the nature of each task.

Each specialised agent—such as those for navigation, data exploration, failure analysis (FMEA), and ship systems—has access to tailored tools and APIs required for task execution. Once an agent completes its task, its output is returned to the \textit{Re-router}, which assesses the response in the context of the original goal. If additional actions are needed, the \textit{Re-router} invokes further agents or returns the final result.

This architecture supports sequential and iterative agent interactions, enabling the system to handle complex, multi-step workflows with minimal human intervention. By encapsulating functionality into modular agents and enabling flexible coordination, the framework offers a scalable and extensible approach to digital twin interactions in maritime operations.

\begin{figure}[h!]
    \centering
    \includegraphics[ width=\linewidth]{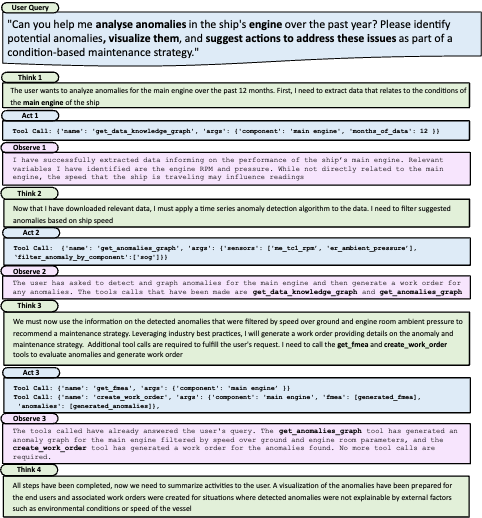}
    \caption{Sample trajectory illustrating the step-by-step process of responding to a user query regarding anomalies in a ship engine. It demonstrates the integration of multiple tools—such as data retrieval, anomaly detection, and failure mode analysis—to detect, visualize, and recommend maintenance strategies. The process culminates in generating a work order based on detected anomalies, filtered by external factors like speed and ambient pressure, ensuring informed decision-making for condition-based maintenance.
}
    \label{fig:agenttrajectory}
\end{figure}

The entire workflow of extracting pertinent sensor data, deploying anomaly detection models, diagnosing root causes is automated by the AI agent \citep{timms2024agentic}. It is insightful to view the trajectory or reasoning process adopted by the LLM to resolve the user query. Figure \ref{fig:agenttrajectory} presents the trajectory including task decomposition, tool calling, and natural language understanding to resolve the task. 

Agent capabilities include: accurately identifying relevant IoT variables associated with asset components (guided by a knowledge graph), detecting exogenous variables that may influence system dynamics (leveraging pre-trained knowledge from an LLM), applying off-the-shelf anomaly detection algorithms to these variables, and presenting the results to the user through visualisations or natural language summaries.

The agent planner module processes the user query and decomposes it into a series of sub tasks. It identifies pertinent tool calls and the associated arguments. For example, it identifies that it needs to extract data related to the main engine so it correctly calls the tool \textit{get\_data\_knowledge\_graph} with the appropriate arguments  for \textit{component} and, \textit{months\_of\_data}.

The agent proceeds through four iterations of think-act-observe (TAO in AI agents’ parlance), to accomplish required subtasks such as data retrieval, applying anomaly detection model, getting failure modes and effects for the component, generating a visualisation and work order, and finally returning all the information to the user. This highlights the powerful capabilities of AI agents in terms of automation and user interaction in natural language. 

\subsection*{Discussion and Conclusion}

Decarbonising the shipping industry is a critical societal goal, with efficiency improvements through digital twin technologies playing a key role in achieving it. However, the inherent complexity of ship systems, vessel heterogeneity, and the diverse range of operational challenges make it difficult to scale digital twin solutions, often requiring significant in-house data science expertise. Agentic LLMs, when augmented with external tools, can significantly accelerate progress towards these objectives by automating and streamlining decision-making processes across various ship types and operational scenarios, making digital twin capabilities more scalable and accessible through a generic natural language interface.

This case study presented an agentic LLM-based framework for anomaly detection in shipping, leveraging real-time data integration, knowledge graphs, and external tools to support complex decision-making in industrial asset management. By evaluating  on real-world data, we demonstrated the ability to streamline the anomaly detection process, and provide contextual insights through dynamic tool selection. This approach offers a scalable solution to the challenges of digitalisation in the maritime sector, addressing issues such as machinery degradation, data integration, and operational efficiency. Future work will focus on enhancing explainability, improving tool orchestration, and integrating additional predictive analytics to further support decarbonisation and automation efforts in shipping.

\section*{Summary of ML Techniques in Aquaculture}

To consolidate the chapter's discussion of machine learning methods, Table~\ref{tab:ml_summary} provides a structured overview of key ML techniques currently applied or emerging in aquaculture. The table highlights representative use cases, the primary benefits these models provide in operational or research settings, and estimated Technology Readiness Levels (TRLs) based on reported deployments and commercial maturity.

The techniques included span both traditional and deep learning models, along with recent advances such as GNNs, LLMs, and AI agents. Additionally, the table includes foundational approaches in time-series modelling, unsupervised learning, and Bayesian inference—methods particularly relevant for data-scarce environments, uncertain decision contexts, or real-time predictive tasks.

By summarising these methods side by side, the table serves as a reference point for selecting appropriate ML approaches depending on the complexity of the aquaculture system, data availability, and deployment goals.

\begin{table}[ht]
\centering
\caption{Key machine learning techniques applied in aquaculture, including typical use cases, benefits, and estimated TRLs based on current research and deployments.}
\begin{tabularx}{\textwidth}{@{}p{2.8cm}|>{\hsize=1.3\hsize}X|>{\hsize=1.3\hsize}X|>{\hsize=.4\hsize}X@{}}
\toprule
\textbf{ML Technique} & \textbf{Application in Aquaculture} & \textbf{Benefits} & \textbf{TRL} \\
\midrule
CNNs & Fish behavior and health monitoring via computer vision & Non-invasive, continuous visual monitoring & 6--8 \\
GNNs & Modeling interactions among environmental and biological systems & Captures system connectivity and spatial relationships & 2--4 \\
SVR & Forecasting dissolved oxygen using satellite or sensor data & Site-specific, accurate environmental predictions & 7--8 \\
LLMs & Natural language interfaces, log analysis, document summarization & Scalable knowledge extraction across diverse sources & 2--4 \\
AI Agents & Autonomous task execution and adaptive response agents & Decision automation and reduced manual intervention & 2--4 \\
Random Forests / Decision Trees & Classification and anomaly detection in water quality or sensor data & Robust predictions with explainable outputs & 7--9 \\
RNNs / LSTMs & Time-series forecasting of DO, temperature, or feeding needs & Handles temporal dependencies in aquaculture processes & 5--7 \\
Unsupervised Learning (e.g., Clustering) & Anomaly detection, behavior clustering, farm typologies & Extracts structure without needing labeled data & 4--6 \\
Bayesian Models & Uncertainty-aware forecasting in small or sparse datasets & Probabilistic modeling and uncertainty quantification & 3--5 \\
\bottomrule
\addlinespace[0.5em]
\multicolumn{4}{@{}l@{}}{\footnotesize \parbox{\textwidth}{\textit{Note:} Technology Readiness Levels (TRLs) are indicative estimates based on literature, commercial trials, and reported deployments.}}
\end{tabularx}
\label{tab:ml_summary}
\end{table}

\section{Conclusions}

Machine learning is emerging as a key enabler in the digital transformation of aquaculture. From predictive analytics and real-time decision support to autonomous monitoring and control, ML technologies can enhance productivity, sustainability, and resilience across the value chain. However, realising this potential depends on improved data infrastructure, model interpretability, and collaboration between AI developers and aquaculture practitioners.

As aquaculture scales to meet global food demands, ML applications provide critical insights, from optimising feeding schedules and monitoring water quality to predicting potential environmental impacts. Emerging technologies such as Graph Neural Networks (GNNs), large language models (LLMs), and AI agents are driving this shift, enhancing traditional aquaculture practices with data-driven intelligence and paving the way for the digital twin's role in real-time, predictive management.

The use of computer vision, for instance, enables continuous and non-invasive monitoring of fish behaviour and health, providing an unprecedented level of observation that supports better decision-making and aligns with sustainability goals. At the same time, GNNs offer promising capabilities for modelling complex, interconnected systems, such as the relationships between water quality parameters, fish behaviour, and external environmental factors, which are often difficult to capture using conventional data analysis techniques.

LLMs and AI agents bring adaptability and scalability to the aquaculture industry. With their broad applicability across diverse datasets and environments, LLMs or foundation models allow for generalization, enabling aquaculture operations to adopt sophisticated AI solutions with minimal fine-tuning. AI agents, on the other hand, facilitate autonomous decision-making processes, potentially transforming daily operations and reducing dependency on manual interventions.

As aquaculture embraces the digital twin concept, the role of AI will expand further into predictive analytics, real-time decision support, and complex ecosystem modelling, driving both operational efficiency and environmental responsibility. Ultimately, the convergence of machine learning and digital twin technology presents a robust framework for achieving sustainable, intelligent aquaculture practices capable of meeting the needs of a growing population while preserving aquatic ecosystems for future generations.

\section*{Acknowledgements}
The research presented in this chapter was supported by multiple sources of funding and institutional support. The work received funding from the European Union’s Horizon Europe Research and Innovation programme under grant agreement no.~101157369.

The authors of Case Study~1 acknowledge project support and computing resources provided by DeepSense, as well as funding from the Mitacs Accelerate program under grant IT33656.

Christopher Whidden is supported by a Natural Sciences and Engineering Research Council of Canada (NSERC) Discovery Grant (RGPIN-2021-02988).

The authors of Case Study~3 gratefully acknowledge support from the Horizon Europe programme under grant agreement no.~101056799.

\bibliography{library}
\bibliographystyle{plainnat}
\end{document}